\documentclass[11pt]{article}

\usepackage[preprint]{acl}

\usepackage{times}
\usepackage{latexsym}
\usepackage[T1]{fontenc}

\usepackage[utf8]{inputenc}

\usepackage{microtype}

\usepackage{inconsolata}

\usepackage{graphicx}
\usepackage{booktabs}
\usepackage{listings}
\usepackage{xcolor}
\usepackage{tcolorbox}
\usepackage{ragged2e}
\usepackage{amsmath, amsthm, amssymb, amsfonts}
\usepackage{algorithm}
\usepackage{algpseudocode}
\usepackage{tabularx}
\usepackage{hyperref}
\usepackage{cleveref}

\newtheorem{theorem}{Theorem}
\newtheorem{proposition}[theorem]{Proposition}

\newtheorem{definition}[theorem]{Definition}

\title{If It's Nice, Do It Twice: We Should Try Iterative Corpus Curation}

\author{Robin Young \\
  Department of Computer Science and Technology \\
  University of Cambridge \\
  Cambridge, UK \\
  \texttt{robin.young@cl.cam.ac.uk}}

\begin{document}
\maketitle

\begin{abstract}
Recent work demonstrates that filtering harmful content from pretraining data improves model safety without degrading capabilities. We propose a natural extension: do it again. A model trained on filtered data can filter the corpus further; training on this cleaner corpus produces an even cleaner model. We provide theoretical analysis showing this process converges to a self-consistent corpus where the model trained on it approves of its own training data. Even under the weak assumption of constant filter quality, iteration yields decay in harmful content. We argue this framework offers a novel form of scalable oversight. While model internals are opaque, the resulting corpus is human-auditable. Even a single iteration produces a large-scale preference annotations over documents, potentially valuable for interpretability research. We derive bounds on capability-safety tradeoffs and outline open questions. We call on researchers with pretraining infrastructure to empirically test this approach.
\end{abstract}

\section{Introduction}

Post-hoc alignment methods are fragile. Models trained on uncurated internet data learn harmful patterns that prove difficult to remove through fine-tuning alone. Jailbreaks routinely circumvent safety training \citep{wei2024jailbroken, perez2022redteaming}, and recent work shows that even benign fine-tuning can erode alignment \citep{qi2023finetuning}. The fundamental problem: once harmful representations are embedded during pretraining, surface-level interventions provide only brittle protection.

A recent line of work addresses safety at its source: pretraining data curation. \citet{anthropic2025filtering} demonstrated that filtering CBRN-related content from pretraining data reduces harmful capabilities by 33\% while preserving general performance. \citet{eleutherai2025deep} showed that such filtering creates ``tamper-resistant'' safeguards that survive adversarial fine-tuning. The intuition is simple: ``even a fully jailbroken model is unlikely to be helpful if it is entirely ignorant of dangerous knowledge.''

These results raise a natural question that, to our knowledge, remains unexplored. \emph{What if we iterate?} A model trained on filtered data should have cleaner representations. This cleaner model might then filter the corpus more effectively, producing an even cleaner training set for the next iteration.

In this paper, we propose the position of iterative corpus curation as a method for building alignment into pretraining. The idea is simple. If filtering works, iterate. We provide theoretical analysis showing this process converges to a self-consistent fixed point, which is a corpus where the model trained on it approves of its own training data. Even under weak assumptions, iteration yields exponential decay in harmful content.

We argue this may offer a novel form of scalable oversight. While model internals are opaque, the resulting corpus is human-auditable using standard methodology. We also observe that iterative curation would produce preference annotations over documents, thus revealing what models believe they should learn from, rather than what they should output.

\section{Related Work}

Recent work has demonstrated the effectiveness of filtering harmful content from pretraining data. \citet{anthropic2025filtering} removed CBRN-related content and found reduced harmful capabilities with no degradation on standard benchmarks. \citet{eleutherai2025deep} showed that such filtering resists adversarial fine-tuning, outperforming post-training safety methods by over an order of magnitude. \citet{maini2025safety} combined filtering with synthetic recontextualization and native refusal training. This line of work responds to findings that post-hoc alignment is fragile; even benign fine-tuning can erode safety training \citep{qi2023finetuning, wei2024jailbroken}.

A related line of work studies models trained iteratively on their own outputs. \citet{ferbach2024selfconsuming} analyze self-consuming generative models, showing convergence to reward-maximizing distributions under curation. \citet{weston2026selfimproving} use post-trained models to improve pretraining data via RL. Constitutional AI \citep{bai2022constitutional} and self-instruction methods \citep{wang2023selfinstruct, li2024selfAlignment} use model-generated feedback, but focus on post-training rather than data curation.

All existing filtering work performs a single pass of filter once, train once. While \citet{grattafiori2024llama} reportedly used Llama 2 to filter training data for Llama 3, this cross-generation iteration has not been systematically studied. No work examines deliberate multi-iteration within a model lineage, and no theoretical analysis of convergence properties exists. We address both gaps by proposing iteration as a deliberate strategy and providing theoretical grounding for why it may work.

\section{Iterative Constitutional Corpus Curation}
\label{sec:method}

Let $D$ be an initial corpus, $\phi$ a constitution specifying what content is acceptable, and $\tau$ a filtering threshold. We define:

\begin{algorithm}[h]
\caption{Iterative Corpus Curation}
\begin{algorithmic}[1]
\State $C_0 \gets D$
\For{$n = 0$ to $N-1$}
    \State $M_n \gets \textsc{Train}(C_n)$
    \State $C_{n+1} \gets \{d \in C_n : \textsc{Score}(M_n, d, \phi) < \tau\}$
\EndFor
\State \Return $C_N, M_N$
\end{algorithmic}
\end{algorithm}

The idea is simple: train a model, use it to filter the corpus, train a new model on the filtered corpus, evaluate, repeat.

\subsection{Intuition}

Consider what each iteration accomplishes:

\textit{Iteration 1:} $M_0$, trained on raw internet data, catches obvious harmful content like explicit slurs, direct instructions for violence, unambiguous toxicity.

\textit{Iteration 2:} $M_1$, trained on the filtered corpus, may catch subtler content. Having seen less toxic content during training, it may have cleaner representations that better distinguish borderline cases.

\textit{Iteration 3+:} Each subsequent model, trained on progressively cleaner data, potentially develops better judgment about what constitutes harmful content.

Much harmful content has approximately zero mutual information with useful knowledge. Racist rhetoric, conspiracy theories, gore descriptions, and ideological manifestos contribute little to a model's ability to perform useful tasks.

This content can be filtered approximately ``for free'' and removed with no capability cost. Early iterations should yield rapid safety improvements by eliminating low-MI unsafe data. We suspect only later iterations would encounter weighing the genuinely dual-use content (chemistry enabling both legitimate research and weapons synthesis) where safety-capability tradeoffs become real.

\section{Theoretical Analysis}
\label{sec:theory}

We analyze convergence under assumptions.

\subsection{Convergence}

\begin{theorem}[Convergence]
The sequence $C_n$ converges to a fixed point $C^*$ in at most $|D|$ iterations.
\end{theorem}

\begin{proof}
Since filtering only removes documents, $C_{n+1} \subseteq C_n$ for all $n$. This is a monotone decreasing sequence in a finite set, hence converges.
\end{proof}

This result requires no assumptions about the filter or training process---only that filtering does not add documents.

\subsection{Fixed Point Characterization}

\begin{definition}[Self-Consistent Corpus]
A corpus $C^*$ is self-consistent with respect to constitution $\phi$ and threshold $\tau$ if:
\[
\forall d \in C^* : \textsc{Score}(\textsc{Train}(C^*), d, \phi) < \tau
\]
\end{definition}

The fixed point is the largest self-consistent corpus reachable from $D$: a model trained on $C^*$ approves of everything in $C^*$.

\subsection{Exponential Improvement}

Even if filter quality remains constant across iterations, harmful content decays exponentially:

\begin{proposition}
Suppose each iteration removes fraction $p$ of remaining harmful content. After $n$ iterations, the fraction of harmful content remaining is $(1-p)^n$.
\end{proposition}

This means we do not need to assume that models improve at filtering. Constant-quality filtering, applied iteratively, still produces exponential decay. If filter quality improves with cleaner training data (an empirically plausible but untested hypothesis), convergence is faster.

\subsection{Capability-Safety Tradeoff}

Let $H \subset D$ be harmful documents, $U \subset D$ be useful documents, and $B = H \cap U$ be dual-use documents containing both harmful and useful content.

Define:
\begin{align}
S(C) &= \frac{|H \setminus C|}{|H|} & \text{(safety: frac harmful removed)} \\
K(C) &= \frac{|U \cap C|}{|U|} & \text{(capability: frac useful retained)}
\end{align}

\begin{theorem}[Capability Bound]
For any corpus $C \subseteq D$:
\[
K(C) \geq 1 - S(C) \cdot \frac{|B|}{|U|}
\]
\end{theorem}

\begin{proof}
To achieve safety $S$, we remove at least $S \cdot |H|$ harmful documents. In the worst case, removed documents are maximally useful (all from $B$). The fraction of useful documents lost is at most $S \cdot |H| / |U| \cdot (|B|/|H|) = S \cdot |B|/|U|$.
\end{proof}

Capability loss is bounded by safety gain times the ``entanglement ratio'' $|B|/|U|$. When harmful and useful content are mostly disjoint (low $|B|$), high safety is achievable with minimal capability loss.

\section{Extension: Preference-Based Curation}
\label{sec:rlaif}

The binary filtering framework admits a natural generalization. Rather than removing documents entirely, we can reweight them based on pairwise preferences, thus connecting iterative curation to the RLHF literature \citep{christiano2017rlhf}.

In standard RLAIF \citep{lee2023rlaif}, a model compares two generations and indicates which is preferable. We can apply the same structure to documents:

\begin{quote}
\emph{Given constitution $\phi$, which document better exemplifies content the model should learn from: $d_1$ or $d_2$?}
\end{quote}

This produces a preference distribution rather than a binary keep/remove decision.

Let $p_n: D \to [0,1]$ be a sampling distribution over documents at iteration $n$. Instead of filtering, we reweight:
\[
p_{n+1}(d) \propto p_n(d) \cdot w(M_n, d)
\]
where $w(M_n, d)$ is document $d$'s ``win rate'' under $M_n$'s judgments; which is the fraction of pairwise comparisons it wins.

The fixed point $p^*$ satisfies: all documents in the support have equal win rates under the model trained on $p^*$. No document dominates another; the corpus is in preference equilibrium.

Binary filtering is aggressive. a document is either fully included or fully excluded, creating sharp tradeoffs for dual-use content. Preference-based reweighting may be gentler. Borderline documents receive reduced weight rather than removal; dual-use content can persist at low probability rather than being deleted entirely. The model retains access to the knowledge without it dominating training. This may achieve better Pareto frontiers of high safety with lower capability cost because more genuinely useful content is downweighted rather than lost.

This framing connects to recent work on self-consuming generative models \citep{ferbach2024selfconsuming}, which analyzes iterative retraining with curated synthetic data. They show convergence to reward-maximizing distributions under certain conditions. Our setting differs as we propose to curate existing documents rather than synthetic generations, but the mathematical structure is similar.

The preference-based view also suggests that iterative corpus curation can be understood as offline RLAIF on documents by learning what to train on, rather than what to output.

\section{Scalable Oversight via Corpus Audit}
\label{sec:oversight}

A central challenge in AI safety is verifying that powerful systems behave as intended. Current approaches focus on model properties. Interpretability aims to understand internal representations, but neurons and circuits remain opaque at scale; RLHF trains reward models to capture human preferences, but reward models are themselves opaque; debate has AIs argue about outputs, but verifying arguments is difficult. Iterative corpus curation offers a different target for verification, namely, the corpus itself.

\begin{center}
\begin{tabular}{lll}
\toprule
\textbf{Approach} & \textbf{Verify} & \textbf{Difficulty} \\
\midrule
Interpretability & Neurons & Hard \\
RLHF & Reward model & Hard \\
Debate & AI arguments & Hard \\
Corpus curation & Documents & Easier \\
\bottomrule
\end{tabular}
\end{center}

Documents are human-readable. We know how to audit text corpora. Libraries, publishers, and content moderators do this routinely. Standard sampling and statistical methods can provide guarantees about corpus quality \citep{cochran1977sampling}.

The oversight protocol is straightforward. Run $N$ iterations of curation, then sample documents from the final corpus $C^*$ for human review. If reviewers verify that sampled documents are acceptable, statistical guarantees extend to the full corpus. This is standard audit methodology, applied to training data rather than model outputs.

This inverts the usual problem. Rather than trying to verify what a model learned by probing opaque representations, interpreting circuits, analyzing reward models, we can verify what it was trained on. The former is an open research problem; the latter is a well-posed problem with established methodology.

\section{Interpretability via Curation Trajectories}
\label{sec:annotations}

Iterative curation produces, as a byproduct, a rich dataset of model judgments that may prove valuable for interpretability research.

Each iteration generates scores for every document:

\begin{center}
\resizebox{\columnwidth}{!}{
\begin{tabular}{ccccc}
\toprule
Document & $M_0$ score & $M_1$ score & $M_2$ score & Status \\
\midrule
$d_1$ & 0.2 & 0.1 & 0.1 & kept \\
$d_2$ & 0.8 & --- & --- & removed (iter 1) \\
$d_3$ & 0.4 & 0.6 & --- & removed (iter 2) \\
\bottomrule
\end{tabular}
}
\end{center}

Unlike RLHF preference data (which captures what models should output), this captures what models believe they should learn from. The rejected documents $D \setminus C^*$ directly reveal what the model considers harmful with no probing or interpretability tools required, just read the documents.

Particularly interesting are documents where successive models disagree. Document $d_3$ above was borderline for $M_0$ (score 0.4) but clearly rejected by $M_1$ (score 0.6). What changed? These disagreement cases may reveal how training data affects constitutional interpretation. By examining what $M_1$ learned (from the cleaner corpus) that caused it to reject content $M_0$ found acceptable, we gain insight into how models develop judgment about harmful content.

Across iterations, the effective interpretation of constitution $\phi$ may drift. The written constitution remains fixed, but each model interprets it differently based on its training. Tracking this drift of which documents change status across iterations, and in which direction may provide a window into how constitutional semantics are grounded in training data.

This could inform constitutional AI research more broadly. If small changes in training data produce large changes in constitutional interpretation, the constitution may be underspecified. Stable interpretations across iterations suggest robust constitutional grounding.

Standard interpretability asks: what has the model learned? This is difficult because representations are distributed and opaque. Curation-based interpretability asks: what does the model think it should learn from? This is easier because the answer is a set of documents we can read.

The two are complementary. Curation trajectories reveal model values at the corpus level; mechanistic interpretability reveals how those values are implemented. Together, they may provide a more complete picture than either alone.

\section{Conclusion}

We propose iterative constitutional corpus curation. If filtering pretraining data works, do it again. A preliminary theoretical analysis is straightforward as convergence is guaranteed, fixed points are self-consistent, and capability loss is bounded by content entanglement. The framework extends naturally to preference-based reweighting, connecting to RLHF theory. Beyond safety benefits, iterative curation produces interpretable artifacts in document-level preference trajectories that reveal how models develop constitutional judgment.

The practical implications may be significant as a form of scalable oversight where verification targets human-readable documents rather than opaque model internals. The experiment is cheap relative to typical pretraining research. A medium sized model over five to ten iterations produces a publishable result regardless of outcome. We encourage researchers with appropriate infrastructure to test whether the theory matches practice.

\section*{Limitations}

This is a position paper with theoretical analysis but no empirical validation. Our core results assume that filtering quality is maintained or improved across iterations. While it is empirically plausible that meta-discussion about harmful content remains even when examples are removed, this is not established. If models require exposure to harmful content to recognize it, filter quality could degrade, causing iteration to stall or drift. Characterizing which regime holds is essential future work.

We lack the pretraining infrastructure and expertise to validate our proposals. The contribution is the concept and theoretical analysis; empirical validation must come from others. The experimental setting would be fairly straightforward conceptually but requires resources we do not have.

Fixed point quality depends entirely on constitution quality. A vague or misspecified constitution $\phi$ produces a fixed point $C^*$ reflecting those flaws. We guarantee self-consistency, not alignment with human values. Relatedly, our analysis treats documents independently. Harmful capabilities might emerge from combinations of individually benign content; corpus-level auditing may miss such compositional risks.

A natural concern is whether models require exposure to harmful content to recognize it. We think this is less problematic than it might appear. A model trained on clean data like Wikipedia, textbooks, and scientific papers when shown overtly harmful content, we posit, will pattern-match it as anomalous relative to its training distribution and the distribution shift itself is signal. The ``confusion requires exposure'' argument is more plausible for subtle dual-use content, but such content dominates only in later iterations after obvious harms are removed. The relevant question is not whether clean models are perfect filters, but whether they are at least as good; and for anomaly detection, cleaner priors may help rather than hurt.

Our analysis treats documents independently, but harmful capabilities can emerge from combinations of individually benign content. A chemistry textbook, a hardware catalog, and a logistics tutorial might each pass review while together enabling dangerous synthesis.

This problem may worsen with iteration. Early rounds remove obviously harmful content; what remains after convergence is precisely content that appears benign in isolation. We may be selecting for compositional risks by filtering out legible ones. Corpus-level auditing inherits this blindspot as reviewers see acceptable individual documents while the combinatorial space of interactions remains unexamined.

Augmenting the constitution with relational criteria (``does this document enable harm in combination with other likely content?'') is theoretically possible but requires models to reason about corpus-level interactions during per-document scoring, which is a significantly harder task. We suspect this limitation applies to any document-level filtering approach, but iteration may exacerbate it.

Proposition 3 assumes each iteration removes a constant fraction of harmful content, yielding exponential decay. With reasonable assumptions, early iterations may capture easy cases (explicit toxicity, obvious violations) while later iterations face subtler content where signal is weaker. Returns may diminish. However, this strengthens rather than undermines our core proposal. If the first iteration catches 80\% of harmful content and the second catches 50\% of what remains, two iterations still remove 90\%, which is substantially better than one. The question of when iteration becomes not worth the effort (third? fifth? tenth?) is empirical, but diminishing returns do not argue against trying more than once.

Several theoretical questions remain open: What determines convergence rate? When do diminishing returns make further iteration not worthwhile? How sensitive is $C^*$ to constitution choice? Can tighter capability bounds be derived with structural assumptions? Is the fixed point unique, or do multiple fixed points exist depending on initialization?

Finally, we focus the conceptual proposal exclusively on text corpora. Extension to multimodal data, code, or other modalities may require different approaches. We also do not address how to construct good constitutions or set appropriate thresholds.

\bibliography{custom}

\appendix

\end{document}